\pdfoutput=1

\documentclass[11pt]{article}

\usepackage{EMNLP2022}

\usepackage{times}
\usepackage{latexsym}

\usepackage[T1]{fontenc}

\usepackage[utf8]{inputenc}

\usepackage{microtype}

\usepackage{inconsolata}

\usepackage{amsmath}
\usepackage{amssymb}
\usepackage[geometry]{ifsym}
\usepackage{booktabs}
\usepackage{multirow}
\usepackage{makecell}
\usepackage{graphicx}
\usepackage{caption}
\usepackage{subcaption}
\usepackage{enumitem}
\usepackage{cleveref}
\crefformat{section}{\S#2#1#3} 
\crefformat{subsection}{\S#2#1#3}
\crefformat{subsubsection}{\S#2#1#3}

\graphicspath{ {figs/} }
\DeclareGraphicsExtensions{.png, .pdf}

\newcommand{\ssc}{C_0}
\newcommand{\rsc}{R}
\newcommand{\hsc}{H}

\DeclareMathOperator*{\argmax}{argmax}

\DeclareFontFamily{U}{mathb}{\hyphenchar\font45}
\DeclareFontShape{U}{mathb}{m}{n}%
{<-6> mathb5 %
 <6-7> mathb6
 <7-8> mathb7 %
 <8-9> mathb8 %
 <9-10> mathb9 %
 <10-12> mathb10 %
 <12-> mathb12 }%
 {}
\DeclareSymbolFont{mathb}{U}{mathb}{m}{n}
\DeclareMathSymbol{\sqbullet}{\mathbin}{mathb}{"0D}

\title{Bridging the Training-Inference Gap for Dense Phrase Retrieval}

\author{
  Gyuwan Kim$^1$ \quad Jinhyuk Lee$^2$\thanks{~~JL currently works at Google Research.} \quad Barlas Oğuz$^3$ \quad Wenhan Xiong$^3$ \\
  {\bf Yizhe Zhang$^3$}\thanks{~~YZ currently works at Apple.} \quad {\bf Yashar Mehdad$^3$} \quad {\bf William Yang Wang$^1$} \\
  $^1$University of California, Santa Barbara \quad $^2$Korea University \quad $^3$Meta AI \\
  \texttt{gyuwankim@ucsb.edu, jinhyuk\_lee@korea.ac.kr} \\
  \texttt{\{barlaso, xwhan, yizhezhang, mehdad\}@fb.com, william@cs.ucsb.edu}
}

\begin{document}
\maketitle
\begin{abstract}
Building dense retrievers requires a series of standard procedures, including training and validating neural models and creating indexes for efficient search.
However, these procedures are often misaligned in that training objectives do not exactly reflect the retrieval scenario at inference time.
In this paper, we explore how the gap between training and inference in dense retrieval can be reduced, focusing on dense phrase retrieval~\citep{lee2021learning} where billions of representations are indexed at inference.
Since validating every dense retriever with a large-scale index is practically infeasible, we propose an efficient way of validating dense retrievers using a small subset of the entire corpus.
This allows us to validate various training strategies including unifying contrastive loss terms and using hard negatives for phrase retrieval, which largely reduces the training-inference discrepancy.
As a result, we improve top-1 phrase retrieval accuracy by \mbox{2$\sim$3} points and top-20 passage retrieval accuracy by 2$\sim$4 points for open-domain question answering.
Our work urges modeling dense retrievers with careful consideration of training and inference via efficient validation while advancing phrase retrieval as a general solution for dense retrieval.

\end{abstract}

\section{Introduction}
Dense retrieval aims to learn effective representations of queries and documents by making representations of relevant query-document pairs to be similar~\cite{chopra2005learning, van2018representation}.
With the success of dense passage retrieval for open-domain question answering (QA)~\citep{lee2019latent,karpukhin2020dense}, recent studies build an index for a finer granularity such as dense \textit{phrase} retrieval~\citep{lee2021learning}, which largely improves the computational efficiency of open-domain QA by replacing the retriever-reader model~\citep{chen2017reading} with a retriever-only model~\citep{seo2019real,lewis2021paq}.
Also, phrase retrieval provides a unifying solution for multi-granularity retrieval ranging from open-domain QA (formulated as retrieving phrases) to document retrieval~\cite{lee2021phrase}, which makes it particularly attractive.

\begin{figure*}[t]
\centering

\includegraphics[width=\textwidth]{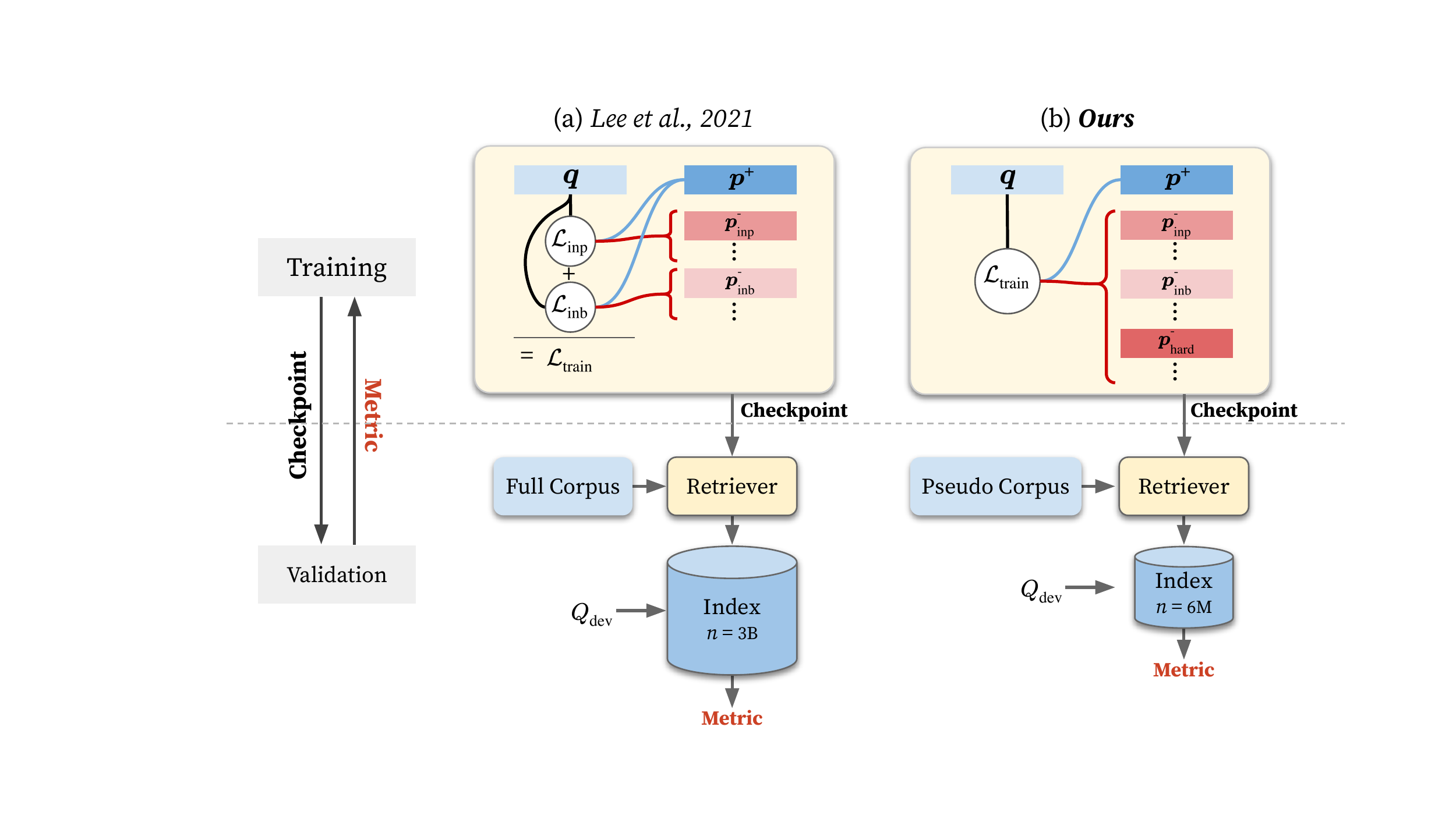}

\caption{
Comparison of the (a) original~\citep{lee2021learning} and (b) proposed procedure for DensePhrases training (top) and validation (bottom).
We unify training loss terms $\mathcal{L}_\text{inp}$ and $\mathcal{L}_\text{inb}$ that enforce the representation of a question ($q$) similar to the representation of a positive phrase ($p^\text{+}$) while contrasting from representations of in-passage negative phrases ($p^\text{-}_\text{inp}$) and in-batch negative phrases ($p^\text{-}_\text{inb}$) respectively into a single term $\mathcal{L}_\text{train}$ and expand negatives in number and difficulty with hard negatives ($p^\text{-}_\text{hard}$).
Also, we use a retrieval accuracy on the development set $Q_\text{dev}$ using a smaller corpus instead of the full corpus as an efficient validation metric for selecting the best checkpoint.
Query-side fine-tuning and token filtering are not described in this overview figure.
}
\label{fig:overview}
\end{figure*}

Building a dense retrieval system involves multiple steps (Figure~\ref{fig:overview}) including training a dual encoder (\textsection\ref{sec:training}), selecting the best model with validation (\textsection\ref{sec:validation}), and constructing an index (often with filtering) for an efficient search  (\textsection\ref{sec:token-filter}).
However, these components are somewhat loosely connected to each other.
For example, model training is not directly optimizing the retrieval performance using the full corpus on which models should be evaluated.
In this paper, we aim to minimize the gap between training and inference of dense retrievers to achieve better retrieval performance.

However, developing a better dense retriever requires validation, which requires building large indexes from a full corpus (e.g., the entire Wikipedia for open-domain QA) for inference with a huge amount of computational resources and time.
To tackle this problem, we first propose an efficient way of validating dense retrievers without building large-scale indexes.
Analysis of using a smaller random corpus with different sizes for the validation reveals that the accuracy from small indexes does not necessarily correlate well with the retrieval accuracy on the full index. 
As an alternative, we construct a compact corpus using a pre-trained dense retriever so that validation on this corpus better correlates well with the retrieval on the full scale while keeping the size of the corpus as small as possible to perform efficient validation.

With our efficient validation, we revisit the training method of dense phrase retrieval~\citep{lee2021learning,lee2021phrase}, a general framework for retrieving different granularities of texts such as phrases, passages, and documents.
We reduce the training-inference discrepancy by unifying previous loss terms to discriminate a gold answer phrase from other negative phrases altogether instead of applying in-passage negatives~\citep{lee2021phrase} and in-batch negatives separately.
To better approximate the retrieval at inference where the number of negatives is extremely large, we use all available negative phrases from training passages to increase the number of negatives and put more weights on negative phrases.
We also leverage model-based hard negatives~\citep{xiong2020approximate} for phrase retrieval, which hasn't been explored in previous studies.
This enables our dense retrievers correct mistakes made at inference time.

Lastly, we study the effect of a representation filter~\citep{seo2018phrase}, an essential component for efficient search.
We separate the training and validation of a phrase filtering module to disentangle the effect of contrastive learning and representation filtering.
This allows us to do careful validation of the representation filter and achieve a better precision/recall trade-off.
Interestingly, we find that a representation filter has a dual role of reducing the index size and also improving retrieval accuracy, meaning smaller indexes are often better than larger ones in terms of accuracy.
This gives a different view of other filtering methods that have been applied in previous studies for efficient open-domain QA~\cite{min2021neurips, izacard2020memory, fajcik2021pruning, yang2021designing}.

We reemphasize that phrase retrieval is an attractive solution for open-domain question answering compared to other retriever-reader models, considering both accuracy and efficiency.
Our contributions are summarized as follows:
\begin{itemize}[noitemsep,topsep=0pt,parsep=0pt,partopsep=0pt]
    \item We introduce an efficient method of validating dense retrievers to confirm and accelerate better modeling of dense retrievers.
    \item Based on our efficient validation, we improve dense phrase retrieval models with modified training objectives and hard negatives.
    \item Consequently, we achieve the state-of-the-art phrase retrieval accuracy for open-domain QA and also largely improve passage retrieval accuracy on Natural Questions~\citep{kwiatkowski2019natural} and TriviaQA~\citep{joshi2017triviaqa}.
\end{itemize}

\section{Related Work}
\label{sec:densephrases}

\paragraph{Dense retrieval}
Retrieving relevant documents for a query~\cite{mitra2018introduction} is crucial in many NLP applications like open-domain question answering and knowledge-intensive tasks~\cite{petroni2021kilt}.
Dense retrievers typically build a search index for all documents by pre-computing the dense representations of documents using an encoder.
Off-the-shelf libraries for a maximum inner product search (MIPS)~\cite{johnson2019billion, guo2020accelerating} enable model training and indexing to be developed independently~\cite{lin2022proposed}.
However, both training dense retrievers and building indexes should take into account the final retrieval accuracy.
In this respect, we aim to close the gap between training and inference of dense retrievers.

\paragraph{Phrase retrieval}
Phrase retrieval~\cite{seo2019real} directly finds an answer with MIPS from an index of contextualized phrase vectors.
This removes the need to run an expensive reader for open-domain QA.  
As a result, phrase retrieval allows real-time search tens of times faster than retriever-reader approaches as an alternative for open-domain QA.
DensePhrases~\citep{lee2021learning} removes the requirement of sparse features and significantly improves the accuracy from previous phrase retrieval methods~\cite{seo2019real,lee2020contextualized}.
\citet{lee2021phrase} show how retrieving phrases could be translated into retrieving larger units of texts like a sentence, passage, or document, making phrase retrieval a general framework for retrieval.
Despite these advantages, phrase retrieval requires building a large index from billions of representations.
In this work, we focus on improving phrase retrieval with more efficient validation.

\paragraph{Validation of dense retrieval}
Careful validation is essential for developing machine learning models to find a better configuration~\citep{melis2018state} or avoid falling to a wrong conclusion.
However, many works on dense retrieval do not clearly state the validation strategy, and most of them presumably perform validation on the entire corpus.
It is doable but quite expensive\footnote{For example, dense passage retrieval (DPR)~\citep{karpukhin2020dense} takes 8.8 hours on 8 GPUs to compute 21-million passage embeddings and 8.5 hours to build a FAISS index.
Also, ColBERT~\citep{khattab2020colbert} takes 3 hours to index 9M passages in the MS MARCO dataset~\citep{nguyen2016ms} using 4 GPUs.} to perform frequent validation and comprehensive tuning.
Hence, it motivates us to devise efficient validation for dense retrieval.
Like ours, \citet{hofstatter2021efficiently} construct a small validation set by sampling queries and using a baseline model for approximate dense passage retrieval but limited to early stopping.
\citet{liu2021can} demonstrate that small and synthetic benchmarks can recapitulate innovation of question answering models on SQuAD~\citep{rajpurkar2016squad} by measuring the concurrence of accuracy between benchmarks.
We share the intuition that smaller and well-curated datasets may lead to the same (or sometimes better) model development while faster but with more focus on the validation process.

\paragraph{Hard examples}
Adversarial data collection by an iterative model (or human) in the loop process aims to evaluate or reinforce models' weaknesses, including the robustness to adversarial attacks~\cite{kaushik2021efficacy,bartolo2021improving,nie2020adversarial,kiela2021dynabench}.
In this work, we construct a compact corpus from a pre-trained dense retriever for efficient validation.
Also, we extract hard negatives from retrieval results of the previous model for better dense representations.

\section{Efficient Validation of Phrase Retrieval}
\label{sec:validation}
Our goal is to train a dense retriever $\mathcal{M}$ that can accurately find a correct answer in the entire corpus $C$ (in our case, Wikipedia).
Careful validation is necessary to confirm whether new training methods are truly effective.
It also helps finding optimal configurations induced by those techniques.
However, building a large-scale index for every model makes the model development process slow and also requires huge memory.  Thus, an efficient validation method could expedite modeling innovations in the correct directions.
It could also allow frequent comparison of different checkpoints when updating a full index simultaneously during the training is computationally infeasible.\footnote{Although some works~\cite{guu2020realm, xiong2020approximate} do asynchronous updates per specific number of training steps and use the intermediate index for better modeling, it requires a huge amount of computational resource.}

Measuring the retrieval accuracy on an index from a smaller subset of the full corpus (denoted as $C^{\star}$) for model validation would be a practical choice, hoping $\argmax_{\mathcal{M} \in \Omega} \text{acc}(D | \mathcal{M}, C^\star) \approx \argmax_{\mathcal{M} \in \Omega} \text{acc}(D | \mathcal{M}, C)$ where $\Omega$ is a set of model candidates and $\text{acc}$ means the retrieval accuracy on a QA dataset $D$.
We first examine how a relative order of accuracy between modeling approaches may change with varying sizes of the random subcorpus (\textsection\ref{sec:random-pseudo-corpus}) and then develop a clever way to construct a compact subcorpus that maintains reasonable correlation with the retrieval accuracy in the full scale (\textsection\ref{sec:hard-pseudo-corpus}).

\subsection{Random Subcorpus}
\label{sec:random-pseudo-corpus}
Reading comprehension (RC) can be regarded a special case of open-domain QA, where a corpus contains only a single gold passage (i.e., $C_q = \{c\}$) for each question.
Here, the subcorpus is question-dependent.
We first gather all gold passages from the development set as a small corpus $\ssc$, a minimal set that contains answers to all development set questions.
We consider a corpus $\rsc_r$ whose size is $r$ times the size of the full corpus by simply appending $\ssc$ with random passages by sampling from the full corpus $C$, i.e., $\ssc \subset \rsc_r \subset C$ and $|\rsc_r| = r|C|$.
We specifically use $r = 1/100, 1/10$ in our experiments.
As the corpus size increases, finding the correct answer from a larger number of possible candidates becomes more difficult, so the retrieval accuracy generally decreases~\citep{reimers2021curse}.

\begin{figure*}[t]
\begin{equation}
\label{eqn:separate_loss}
\mathcal{L}_\text{train}^\text{org} = -\log \frac{e^{s(q, p^\text{+}; c)}}{e^{s(q, p^\text{+}; c)} + \sum_{(p^\text{-}; c) \in N_\text{inp}} e^{s(q, p^\text{-}; c)}}
- \lambda \log \frac{e^{s(q, p^\text{+}; c)}}{e^{s(q, p^\text{+}; c)} + \sum_{(p^\text{-}; c') \in N_\text{inb} \cup N_\text{prb} } e^{s(q, p^\text{-}; c')}}
\end{equation}
\begin{equation}
\label{eqn:unified_loss}
\mathcal{L}_\text{train} = -\log \frac{e^{s(q, p^\text{+}; c)}}{e^{s(q, p^\text{+}; c)} + \sum_{(p^\text{-}; c') \in N_\text{inp} \cup N_\text{inb} \cup N_\text{prb} \cup N_\text{hard}} \lambda(p^\text{-}) e^{s(q, p^\text{-}; c')}}
\end{equation}
\end{figure*}

DensePhrases~\citep{lee2021learning} simply choose the best checkpoint with the highest RC accuracy assuming that a model with better RC accuracy leads to a better retrieval accuracy, or use the last checkpoint at the end of the training.\footnote{RC accuracy generally improves during the training of DensePhrases as the training loss directly optimizes it.}
It is problematic since our preliminary experiments demonstrate that the RC accuracy and the retrieval accuracy on different sizes of corpus including the full corpus, do not necessarily correlate well with each other.
Using a large subcorpus is better for accurate validation not to deviate much from the trends of retrieval accuracy of a full corpus.
However, a smaller subcorpus would be better in terms of validation efficiency.
This trade-off drives us to design a better way of constructing a validation corpus.

\subsection{Hard Subcorpus}
\label{sec:hard-pseudo-corpus}
The retrieval accuracy given a subcorpus $C^{\star}$ should have a high correlation with the retrieval accuracy over the full corpus and the size of corpus $|C^{\star}|$ should be small enough (or as small as possible) for efficient validation.
For a reasonably accurate dense retriever, it is relatively easy to discriminate a gold phrase from other phrases in random passages.
Therefore, it is better to collect a subcorpus with \textit{hard} passages to test dense retrievers on a similar condition to a full corpus which includes many difficult phrases to discriminate if the corpus can have a limited number of negative passages.

We construct a hard corpus $\hsc_k$ with a compact size using a pre-trained retriever $\bar{\mathcal{M}}$ to extract all context passages of top-$k$ retrieved phrases for all query $q$ in the development set $Q_\text{dev}$, and $\ssc$ is merged to always include an answer, i.e., $\hsc_k = \ssc \cup \bigcup_{(q, a) \in Q_\text{dev}} \bar{\mathcal{M}}_k(q | C)$ where $\mathcal{M}_k(q | C)$ denotes the top-$k$ passage retrieval results for a query $q$ from the model $\mathcal{M}$.
If $\bar{\mathcal{M}}$ is reasonably accurate, negative examples retrieved by $\bar{\mathcal{M}}$ will make our new model $\mathcal{M}$ difficult to find a correct answer.  
We expect the retrieval accuracy from $\hsc_k$ quickly drops as $k$ increases and reaches close to the retrieval accuracy on the full corpus $C$ with a manageable $k$ so that we can use retrieval accuracy on a hard subcorpus for efficient validation.
It keeps the relative order of models with a much smaller size than the random subcorpus.

\section{Optimized Training of DensePhrases}
\label{sec:training}

In this section, we briefly review the original training method of DensePhrases (\textsection\ref{sec:densephrases-training}) and improve it further to reduce the gap between training and retrieval in inference by modifying the training objective (\textsection\ref{sec:unified-loss}) and introducing additional training with hard negatives (\textsection\ref{sec:hard-negative}).

\subsection{Background: Training of DensePhrases}
\label{sec:densephrases-training}
The question encoder and the phrase encoder are jointly trained using reading comprehension datasets.
A phrase $p$ is represented as a concatenation of start and end token vectors from the contextualized representations of a context passage $c$ using a phrase encoder.
A question $q$ is represented as a concatenation of vectors using two different encoders for the start and the end.

The main training objective is a sum of the two separate contrastive loss terms weighted by the $\lambda$ coefficient as formally defined in Equation~\ref{eqn:separate_loss}.\footnote{We denote the similarity score between a question $q$ and a phrase $p$ as $s(q, p; c)$. While the score and the loss term of start and end tokens are separately calculated in practice, we abbreviate it in the equation for simplicity.}
One is for contrasting a phrase token of positive start/end position ($p^\text{+}$) to that of other positions in the context passage ($N_\text{inp} = \{ (p; c) \neq (p^\text{+}; c) | p \in c \}$). 
Another is for contrasting the same token to other positive tokens in a current ($N_\text{inb} = \{ (p'; c') \neq (p^\text{+}; c) | (q', p'; c') \in \mathcal{B}\}$) or previous $T$ mini-batches ($N_\text{prb} = \{ (p'; c') | (q', p'; c') \in \mathcal{B}_\text{pre} \}$).
The numbers of negatives are $|N_\text{inp}| = L - 1$, $|N_\text{inb}| = B - 1$, and $|N_\text{prb}| = B \times T$ where $L$ is the sequence length of context passages and $B$ (= $|\mathcal{B}|$) is the batch size.

To learn better representations, the dual encoder is first pre-trained with question-answer pairs generated by a question generation model as a data augmentation mainly for better reading comprehension capability and then fine-tuned with original question-answer pairs.
Also, knowledge distillation~\citep{hinton2015distilling} from a stronger reading comprehension model based on a cross encoder to the dual encoder is performed.
Lastly, the token filtering classifier (explained more in \textsection\ref{sec:token-filter}) that discriminates tokens likely to be a start or end of the answer phrases using a linear classifier on top of phrase representations is jointly trained with the dual encoder.
We omit two additional loss terms from Equation~\ref{eqn:separate_loss} for knowledge distillation and a token filtering classifier loss for brevity.

\paragraph{Query-side Fine-tuning}
Documents in the reading comprehension dataset used for the training take only a tiny portion of the entire Wikipedia, and only a small number of negatives for each question-phrase pair are contrasted compared to billion-scale possible phrase candidates in the test time.
The query encoder can be further fine-tuned to reduce this discrepancy between training and inference while fixing the phrase encoder and the index by maximizing the likelihood of the gold answer among retrieved phrases for each question.
Using more and harder negatives is also an effective way to reduce this gap.

\subsection{Unified Loss}
\label{sec:unified-loss}
The original training objective of DensePhrases (Equation~\ref{eqn:separate_loss}) has separate terms for finding a relevant passage (in/pre-batch negatives) and finding the exact phrase position in the passage (in-passage negatives).
However, we should find an answer phrase among all possible candidates at once during the test time.
Therefore, we modify the loss term as a unified version (Equation~\ref{eqn:unified_loss}) by putting all negatives together into the contrastive targets.

We also introduce the $\lambda$ coefficient to the unified loss to give weights to negatives.
It opens a new question of how we should set the value of $\lambda$.
The role of $\lambda$ can be interpreted in two ways.
First, multiplying $\lambda$ to an exponential of a score is equivalent to adding a positive value to the score ($\lambda e^s = e^{s + \log \lambda}$), and then the loss term becomes the soft version of margin-based loss.
Second, using $\lambda$ can mimic the inference time where the number of negative tokens is much larger by duplicating a negative $\lambda$ times ($\lambda e^s = e^s + e^s + ... + e^s$) to close the gap between training and test. 
Based on the second interpretation, we set different value of $\lambda$ depending on where negative phrase $p^\text{-}$ is from: $\lambda(p^\text{-}) = \lambda_\text{inp} \delta(p^\text{-} \in N_\text{inp}) + \lambda_\text{inb} \delta(p^\text{-} \in N_\text{inb} \cup N_\text{prb}) + \lambda_\text{hard} \delta(p^\text{-} \in N_\text{hard})$.

We extend to use all tokens in context passages with a similar intuition that contrasting with as many tokens as possible could be helpful instead of using only start/end position tokens to in/pre-batch negatives.
It changes in/pre-batch negatives to $N_\text{inb} = \{ (p; c') \neq (p^\text{+}; c) | p \in c', (q', p'; c') \in \mathcal{B} \}$ and $N_\text{prb} = \{ (p; c') | p \in c', (q', p'; c') \in \mathcal{B}_\text{pre} \}$) and their sizes $|N_\text{inb}| = B \times L - 1$ and $|N_\text{prb}| = B \times T \times L$.
This change also increases the number of negatives hundreds of times and turns out empirically advantageous.

\subsection{Hard Negatives for Phrase Retrieval}
\label{sec:hard-negative}
We exploit hard negatives to benefit phrase retrieval, a widely used technique for passage retrieval\footnote{\citet{karpukhin2020dense} use one hard negative obtained from BM25 per example in addition to in-batch negatives for training a dual encoder. \citet{xiong2020approximate} globally select hard negatives from the entire corpus with asynchronously index updates for faster convergence. RocketQA \cite{qu2021rocketqa} denoises hard negatives using cross encoder. The best strategy for hard negative mining and training is still an open problem in dense retrievals.} but never fully examined for phrase retrieval.
We perform a model-based hard negative mining by retrieving top phrases using a pre-trained dual encoder and an index built from this model.
We filter out phrases whose surrounding passage includes a gold answer (\textsection\ref{sec:hard_negative_mining}) and then fine-tune the model with extracted hard negatives  (\textsection\ref{sec:hard_negative_training}).
Although we do it only once, this process could be repeated until convergence.

\begin{figure*}[t]
\centering
\begin{subfigure}[b]{0.32\textwidth}
\centering
\includegraphics[width=\textwidth]{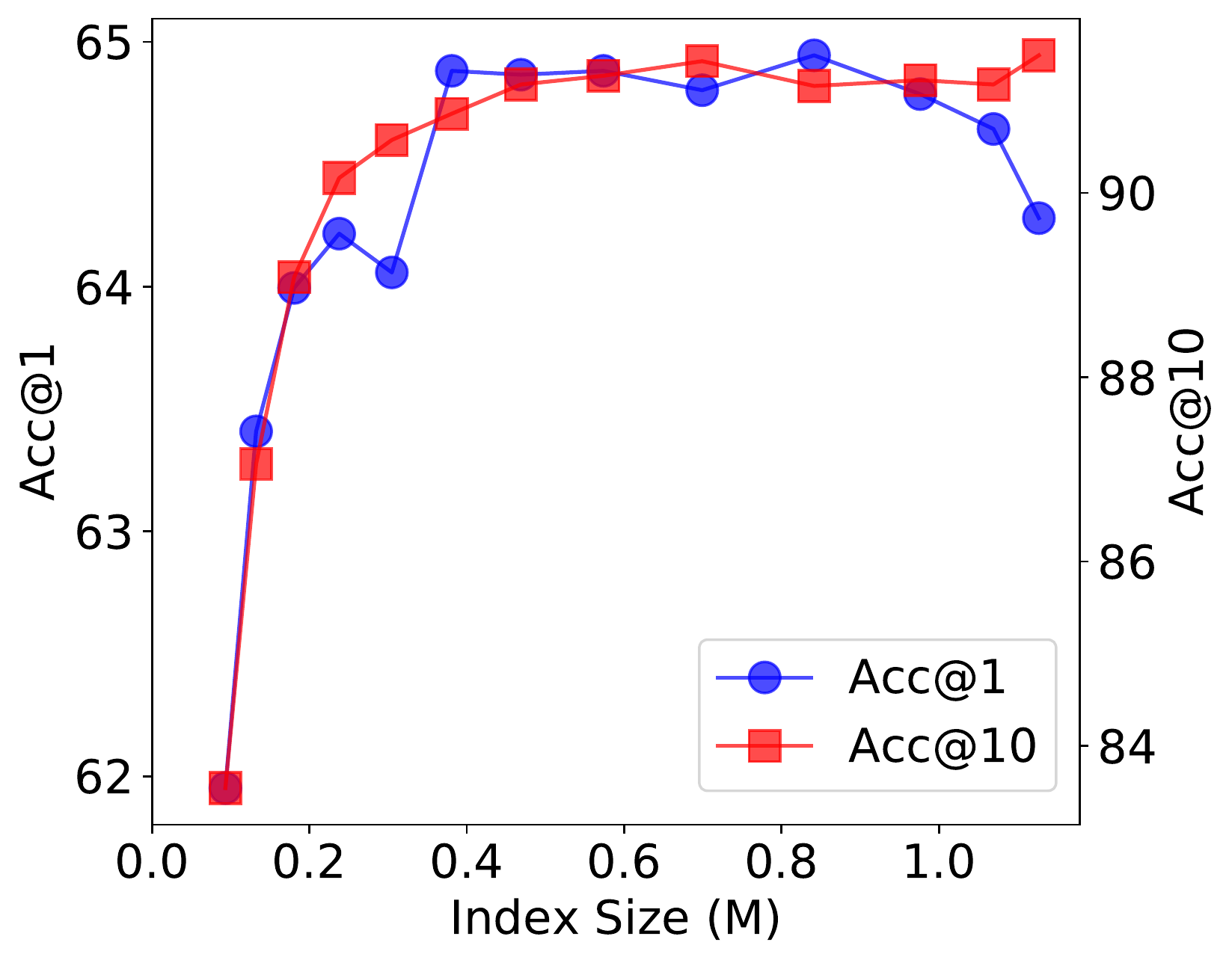}
\caption{$\ssc$}
\label{fig:sod-nq-dev}
\end{subfigure}
\hfill
\begin{subfigure}[b]{0.32\textwidth}
\centering
\includegraphics[width=\textwidth]{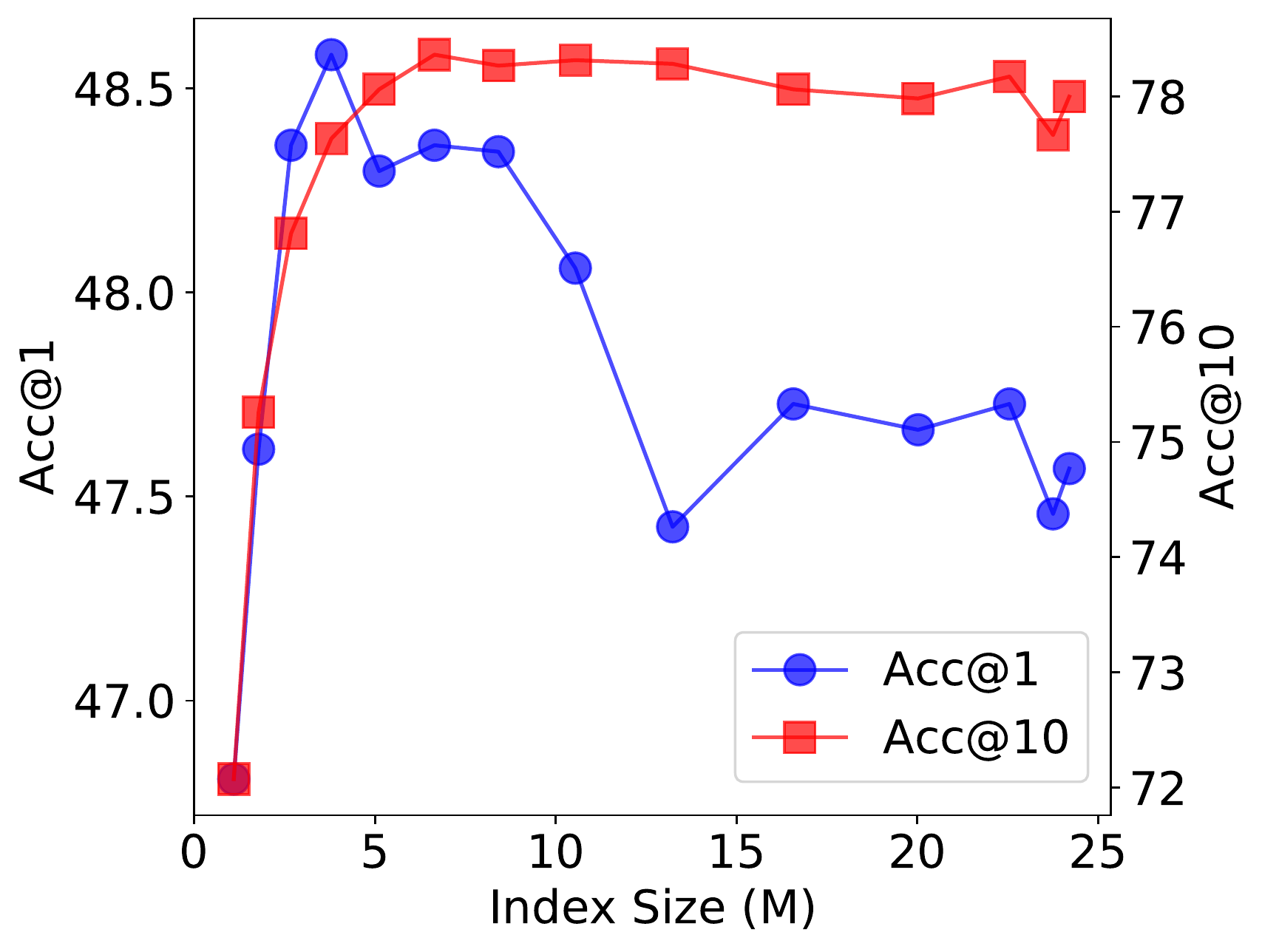}
\caption{$\rsc_{1/100}$}
\label{fig:sod-wiki-dev}
\end{subfigure}
\hfill
\begin{subfigure}[b]{0.32\textwidth}
\centering
\includegraphics[width=\textwidth]{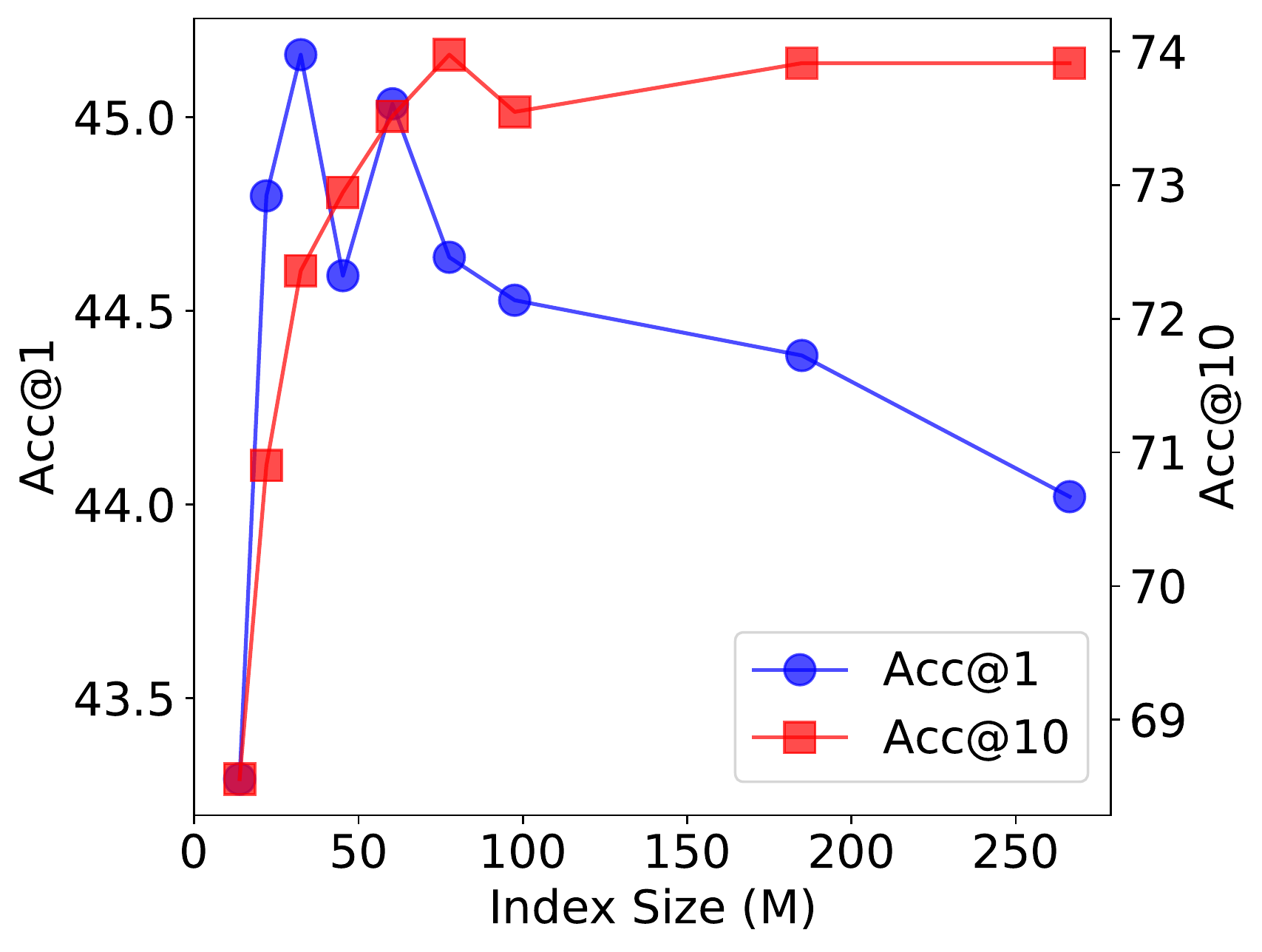}
\caption{$\rsc_{1/10}$}
\label{fig:sod-wiki-dev-noise}
\end{subfigure}

\caption{
The trade-off between the index size and validation retrieval accuracy by changing filter thresholds on random subcorpora with different sizes, (a) $\ssc$, (b) $\rsc_{1/100}$, and (c) $\rsc_{1/10}$.
A threshold that gives better accuracy with a smaller index size exists.
Acc@1 (blue) is more unstable than Acc@10 (red).
Interestingly, the index size of peak EM@1 is smaller than that of peak Acc@10.
}
\label{fig:trade_off}
\end{figure*}

\subsubsection{Hard Negative Mining}
\label{sec:hard_negative_mining}
We use a encoder model and phrase index from the first round to extract model-based hard negatives from top-$k$ phrase retrieval results for questions in the training set.
Using high-quality hard negatives by removing false negatives is important to train a better model.

We exclude examples when a context passage of a retrieved phrase contains an answer.
A context passage corresponding to a retrieved phrase can be restored using information stored along with the index.
It helps to focus more on topically different documents and shares the intuition from the analysis in \citet{lee2021phrase} that DensePhrases less rely on the topical relevance than DPR.
Compared to more strict condition based on exact match that may miss almost correct phrases with a minor error in the boundary by misclassifying the exact position,
it reduces about 20\% of negative pairs, hopefully reducing false negatives and achieving higher accuracy gain.  
Besides, these rules are somewhat loose in that there could be multiple possible answers to a question, and different representations for the same entity could exist since the annotated answer list is imperfect.
We left filtering based on a cross encoder~\citep{qu2021rocketqa, ren2021rocketqav2} to future work due to the convenience of automatic filtering.

\subsubsection{Training with Hard Negatives}
\label{sec:hard_negative_training}
After the hard negative mining, we fine-tune a dual encoder with the hard negatives.
We sample $h$ hard negatives for each training step and append them to negative targets for the loss calculation.\footnote{If the number of hard negatives after removing false negatives is less than $h$, we sample random passages to match the number of hard negatives for parallel computation.}
We expect that hard negatives give a better training signal than random in/pre-batch negatives \cite{xiong2020approximate} because those are examples difficult to discriminate for the previous model.
Moreover, hard negatives extracted from the larger corpus could expose a model to other diverse documents than the original training dataset.
This is similar to query-side fine-tuning but differs in that both the question encoder and phrase encoder are updated.

There are different possible options for choosing $N_\text{hard}$.
First, we may include only corresponding hard negatives for each example or all hard negatives in a mini-batch.
Second, we may include only each negative's start/end position token or all tokens in the context passage.
Similar to \textsection\ref{sec:unified-loss}, we include all tokens in the context passage of all hard negatives in a mini-batch for $N_\text{hard}$.
Using all available tokens is generally better because they potentially belong to the final negative phrase candidates in inference. 
Training with larger numbers of negatives is beneficial to reduce the gap between training and inference.
Including all of them does not induce significant additional memory overhead since we should encode the same number ($h$) of passages regardless of different options. 
Therefore, we use $N_\text{hard} = \bigcup_{(\hat{p}; \hat{c}) \in H(q, p^\text{+}; c), (q, p^\text{+}; c) \in \mathcal{B}} \{(p; \hat{c}) | p \in \hat{c} \} $ as all tokens from all hard negatives in a mini-batch where $H$ is a set of the sampled $h$ hard negatives and $|N_\text{hard}| = B \times L \times h$.

\section{Token Filtering}
\label{sec:token-filter}
Representation filtering is often applied in practice to reduce the index size for efficient search~\citep{min2021neurips}.
For phrase retrieval, tokens that are not likely to be a start/end position of an answer are filtered out using a trained filter classifier based on a logit score for each token to reduce an index size without losing accuracy much.
Only tokens with a score larger than a specific threshold are kept.  
After the filtering, the index is compressed using optimized product quantization~\citep{ge2013optimized}.

\subsection{Token Filter Threshold}
A filter threshold for the token filter determines a trade-off between the index size (efficiency) and retrieval accuracy (Figure~\ref{fig:trade_off}).
Interestingly, we find that token filtering can even improve retrieval accuracy.
As we increase a threshold from a very small value (not filtering), the accuracy fluctuates but generally increases until a specific threshold because the filter successfully reduces the number of candidates, making prediction easier. 
After that threshold, the accuracy drops quickly because the filter starts to leave out many correct tokens. 

However, finding the peak retrieval accuracy requires a manual search of different thresholds after indexing and evaluating.
Since using it as a validation metric is expensive, we first select the best checkpoint based on retrieval accuracy without performing any token filtering.
Especially when the token filter is in the middle of training, the peak threshold will vary, and using a specific fixed threshold would not be fair.
Also, the best threshold changes depending on the corpus size, so choosing a threshold for the full corpus based on a smaller corpus is not straightforward.

\subsection{Token Filter Training and Valiation}
\citet{lee2021learning} jointly train a token filter classifier with a dual encoder.
It is convenient in that an additional training process is not required, while we should tune on the weight for a loss before adding to the overall training loss.
Training pushes phrase vectors into two moving cones toward the start and end vectors since a logit is a dot product score between a phrase vector and a start/end vector.
It has two potential disadvantages: (1) phrase representations are concentrated on a subset of the entire feature space, so the expressiveness of the model is not fully exploited, and (2) optimization is more difficult because of the moving targets.

To address the issues, we train a token filter after training a phrase encoder.
We could expect that the two-stage training process encourages phrase representations to be distributed over the space.
Moreover, we can validate the token filter separately due to the separate training process and pick the best one.
We can not decide the threshold during the filter training, so we use the AUC-PR metric for filter validation by measuring precision and recall by sweeping all threshold values.

\section{Experiments}
\label{sec:experiments}
To show the effectiveness of our proposed method, we evaluate DensePhrases models on open-domain QA benchmarks following the experimental setup of \citet{lee2021learning, lee2021phrase}.\footnote{\url{https://github.com/princeton-nlp/DensePhrases}}

\paragraph{Datasets}
We measure phrase retrieval accuracy and passage retrieval accuracy on two open-domain QA datasets following the standard train/dev/test splits: Natural Questions \cite{kwiatkowski2019natural} and TriviaQA \cite{joshi2017triviaqa}.
We first train our phrase retrieval models on Natural Questions (DensePhrases$^\heartsuit$) or on multiple reading comprehension datasets (DensePhrases$^\spadesuit$), namely Natural Questions, WebQuestions~\citep{berant2013semantic}, CuratedTREC~\citep{baudivs2015modeling}, TriviaQA, and SQuAD~\citep{rajpurkar2016squad}.
Then each model is further query-side fine-tuned on Natural Questions and TriviaQA.
We build the phrase index with smaller subsets of corpora ($R_r$ or $H_k$) for validation and use the 2018-12-20 Wikipedia snapshot ($C$) for the final inference.

\paragraph{Training details}
We use the same training hyperparameters of the original DensePhrases except for the batch size $B=48$.
We set the number of training epochs to 2 with the generated question-answer pairs and increase the number of training epochs to 10 with the standard reading comprehension dataset for more careful validation.
We set $\lambda_\text{inb} = 256$ and $\lambda_\text{inp} =\lambda_\text{hard} = 1$.
We set $k=10$ and $h=1$ for the hard negative mining and sampling.

\paragraph{Token filtering}
Our token filter achieves an improved AUC-PR value over the filter from the original DensePhrases model (e.g., 0.348 vs. 0.307).
We use a filter threshold of -3.0 for the index with the full corpus.
This threshold reduces the index size by more than 70\% of the original size.

\begin{figure}[t]
\centering

\includegraphics[width=\linewidth]{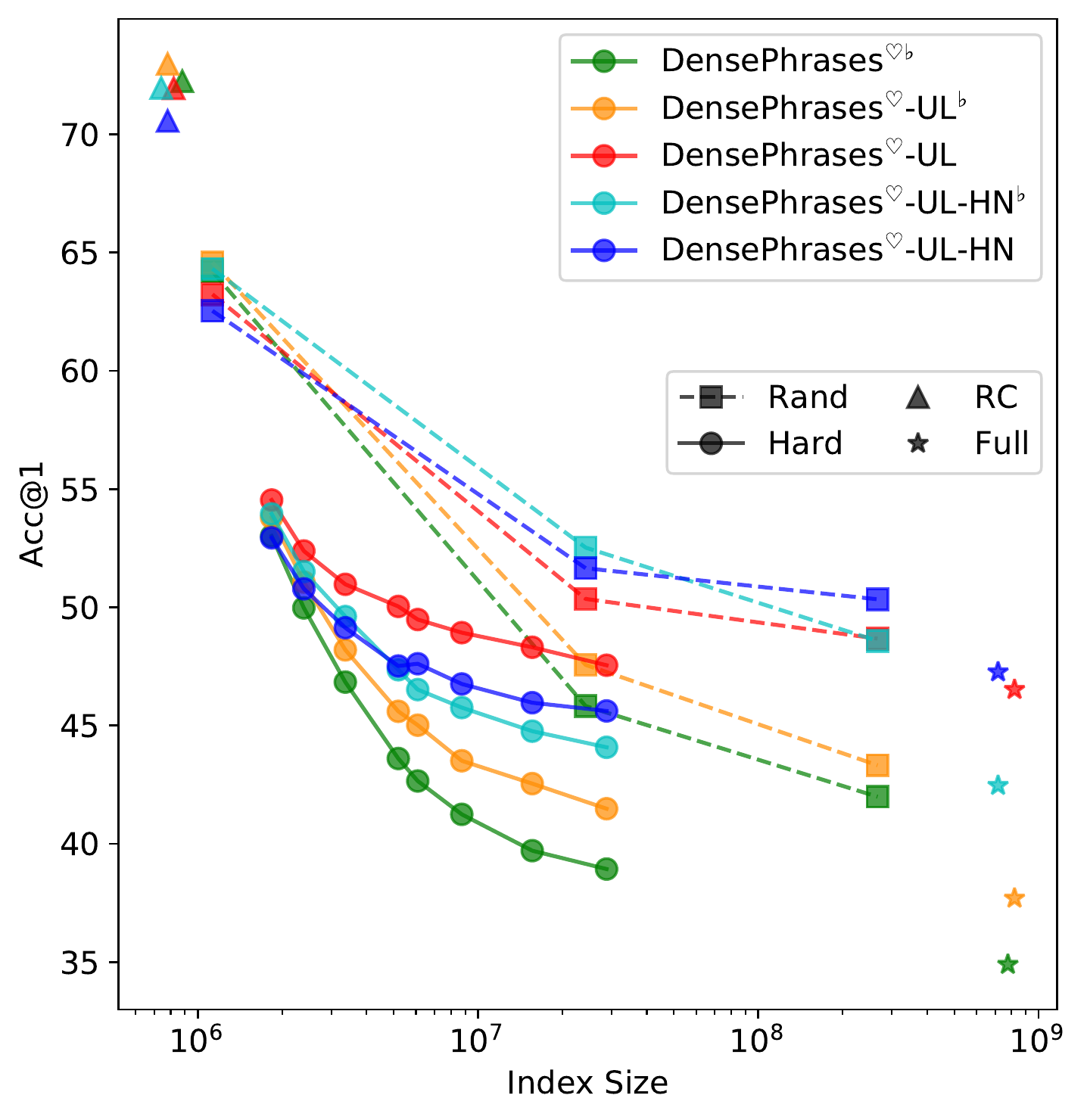}

\caption{
Validation results on open-domain QA.
We plot retrieval accuracy (Acc@1) on indexes with different sizes (log-scale) from random and hard subcorpora.
Random subcorpora ($\sqbullet$) starts with $C_{0}$ and is extended to $R_{1/100}$ and $R_{1/10}$.
Hard subcorpora ($\bullet$) include $H_k$ for $k \in \{1, 2, 4, 8, 10, 16, 32, 64\}$.
We also plot reading comprehension (RC) accuracy\footnotemark~and retrieval accuracy on the full index with filtering.
We compare five different models with or without our proposed training methods and query-side fine-tuning.
All models are trained and evaluated on Natural Questions.
UL, HN, and $^\flat$ indicate a model trained with the unified loss, hard negatives, and before query-side fine-tuning. 
}
\label{fig:validation}
\end{figure}

\footnotetext{Since the length of a passage for each question varies, we put aside points corresponding to RC on the left of the figure with arbitrary small index sizes.}

\subsection{Model Validation}
In our preliminary experiments, we observe that the best checkpoint among training epochs differs depending on the corpus size (especially for small scale). 
Figure~\ref{fig:validation} shows validation retrieval accuracy of the DensePhrases models with different training methods on various sizes of random and hard subcorpora.
The retrieval accuracy on the hard subcorpus rapidly drops and reaches close to the retrieval accuracy on the full corpus as $k$ increases with moderately increasing the index size.
On the other hand, retrieval accuracy on a random subcorpus is higher than on a hard subcorpus with similar index size.
For instance, retrieval accuracies on $H_8$ (5.1M) are lower than those on $\rsc_{1/100}$ (24.2M) with 4 times smaller index, and retrieval accuracies on $\hsc_{16}$ (8.7M) are lower than those on $\rsc_{1/10}$ (266.4M) with 30 times smaller index.
It indicates that a hard subcorpus can effectively imitate inference with a full corpus where correct retrieval is the most difficult.

The relative order of accuracy between models on hard subcorpus converges quickly at around $\hsc_{10}$ (6.1M).
However, the order of accuracy when using random subcorpus changes from $\rsc_{1/100}$ to $\rsc_{1/10}$ showing the difficulty of efficient validation.
On the other hand, retrieval accuracy on a hard subcorpus is more stable and serves as an efficient validation metric. 

Our validation results clearly demonstrate that unified loss is helpful.
Query-side fine-tuning also harms the RC accuracy and the retrieval accuracy with $\ssc$ (1.1M) while improving the retrieval accuracy with larger indexes.
It shows how a wrong conclusion can be made from small-sized corpora.

\begin{table}[t!]

\centering
\small
\def\arraystretch{0.8}
\begin{tabular}{lcc}
\toprule
\multicolumn{1}{l}{\multirow{2}{*}{\textbf{Model}}} & NQ & TQA \\
\cmidrule(lr){2-2} \cmidrule(lr){3-3}
& \scriptsize Acc@1 & \scriptsize Acc@1 \\
\midrule
DPR$^{\diamondsuit}$ + BERT reader      & 41.5 & 56.8 \\
DPR$^{\spadesuit}$ + BERT reader        & 41.5 & 56.8 \\
\midrule
RePAQ$^{\diamondsuit}$ {\scriptsize (retrieval-only)}   & 41.2 & 38.8 \\
RePAQ$^{\spadesuit}$ {\scriptsize (retrieval-only)}     & 41.7 & 41.3 \\
DensePhrases$^{\heartsuit}$             & 40.9 & 50.7 \\
DensePhrases$^{\spadesuit}$             & 41.3 & 53.5 \\
\midrule
DensePhrases$^{\heartsuit}$-UL          & 43.5 & 51.3 \\
DensePhrases$^{\heartsuit}$-UL-HN       & \textbf{44.0} & 47.0 \\
DensePhrases$^{\spadesuit}$-UL        & 42.4 & \textbf{55.5} \\
\bottomrule
\end{tabular}

\caption{
Open-domain QA phrase retrieval test results. 
We report top-1 accuracy (Acc@1).
$^{\diamondsuit}$: trained on each dataset independently. 
$^{\spadesuit}$: trained on multiple datasets.
$^{\heartsuit}$: trained on Natural Questions datasets.
}

\label{tab:phrase-retrieval}
\end{table}

\subsection{Phrase Retrieval}
Table~\ref{tab:phrase-retrieval} summarizes end-to-end open-domain QA results.
Both unified loss and hard negatives are shown to be effective.
With our improved training methods, the best model surpasses the original DensePhrases model by 2.7 points in Natural Questions and 2.0 points in TriviaQA, achieving the state-of-the-art retrieval-only open-domain QA performance.

\begin{table*}[t!]

\centering
\small
\def\arraystretch{0.80}
\begin{tabular}{lcccccccccc}
\toprule
\multicolumn{1}{l}{\multirow{2}{*}{\textbf{Model}}} & \multicolumn{5}{c}{Natural Questions} & \multicolumn{5}{c}{TriviaQA} \\
\cmidrule(lr){2-6} \cmidrule(lr){7-11}
& \scriptsize Acc@1 & \scriptsize Acc@5 & \scriptsize Acc@20 & \scriptsize MRR@20 & \scriptsize P@20 
& \scriptsize Acc@1 & \scriptsize Acc@5 & \scriptsize Acc@20 & \scriptsize MRR@20 & \scriptsize P@20 \\
\midrule
DPR$^{\diamondsuit}$                    & 46.0 & 68.1 & 79.8 & 55.7 & 16.5 & 54.4 &   -  & 79.4 &   -  &   -  \\
DPR$^{\spadesuit}$                      & 44.2 & 66.8 & 79.2 & 54.2 & 17.7 & 54.6 & 70.8 & 79.5 & 61.7 & 30.3 \\
DensePhrases$^{\heartsuit}$             & 50.1 & 69.5 & 79.8 & 58.7 & 20.5 &   -  &   -  &   -  &   -  &   -  \\
DensePhrases$^{\spadesuit}$             & 51.1 & 69.9 & 78.7 & 59.3 & 22.7 & 62.7 & 75.0 & 80.9 & 68.2 & 38.4 \\
\midrule
DensePhrases$^{\heartsuit}$-UL          & 57.1 & 75.7 & 83.7 & 65.2 & 22.0 & 62.0 & 74.6 & 80.6 & 67.6 & 33.3 \\
DensePhrases$^{\heartsuit}$-UL-HN       & \textbf{58.6} & 75.7 & 83.4 & \textbf{66.1} & 21.9 & 60.3 & 73.3 & 79.6 & 66.1 & 32.3 \\
DensePhrases$^{\spadesuit}$-UL          & 56.7 & \textbf{75.9} & \textbf{83.8} & 65.2 & \textbf{23.7} & \textbf{65.0} & \textbf{76.6} & \textbf{82.7} & \textbf{70.2} & \textbf{39.0} \\
\bottomrule
\end{tabular}

\caption{
Open-domain QA passage retrieval test results. 
We report top-$k$ passage retrieval accuracy (Acc@$k$, for $k \in \{1, 5, 20\}$), mean reciprocal rank at 20 (MRR@20), and precision at 20 (P@20).
$^{\diamondsuit}$: trained on each dataset independently. 
$^{\spadesuit}$: trained on multiple datasets.
$^{\heartsuit}$: trained on Natural Questions datasets.
}

\label{tab:passage-retrieval}
\end{table*}

\subsection{Passage Retrieval}
Table~\ref{tab:passage-retrieval} summarizes open-domain QA passage retrieval results.
Our method also improves passage retrieval accuracy significantly.
The best model improves top-20 passage retrieval accuracy by 4.0 points in Natural Questions and 1.8 points in TriviaQA.
It again shows that DensePhrases can be used for passage retrieval as well.
We may use DensePhrases as a building block of other tasks and expect to achieve good phrase retrieval performance with expressive reader models like FiD~\citep{izacard2021leveraging}.

\subsection{Discussion on Hard Negatives}
From \Cref{fig:validation}, we observe that with hard subcorpora, the model trained with hard negatives (cyan) shows higher validation accuracy than the model without hard negative training (yellow) before query-side fine-tuning, but their order changes after query-side fine-tuning (blue vs. red).
This is because hard negative mining process is similar to hard corpus construction, blurring the accurate estimation of validation performance.
However, we pick the best model before the query-side fine-tuning, which lets us to decide to go with hard negatives (due to cyan vs. yellow) and achieve state-of-the-art performance with the full index.

From \Cref{tab:phrase-retrieval}, we observe that hard negatives improve in-domain accuracy but harm the out-of-domain accuracy.
Since hard negative passages are close to the original training data, it improves the performance of questions from the same domain but could cause overfitting and harm the generalization ability.
This observation solicits better hard negative mining methods.

\section{Conclusion}
In this study, we aim to bridge the gap between training and inference of phrase retrieval.
We first develop an efficient validation metric that measures retrieval accuracy on the index from a small corpus with hard passages using a pre-trained retriever.
Based on this validation, we show that the improvements in training of dense phrase retrieval with unified loss and hard negatives are effective.
As a result, we achieve state-of-the-art phrase retrieval and passage retrieval accuracy in open-domain question answering among retrieval-only approaches.

Our work demonstrates that thorough validation is crucial for the accurate and efficient development of phrase retrieval with large corpus.
Also, we prove that modeling and training methods should be designed closely to retrieval in inference time.
Despite its remarkable efficiency and flexibility, phrase retrieval has been relatively less studied than passage retrieval.
We believe that our work can encourage more study on phrase retrieval with an efficient development cycle.
Furthermore, we hope that our findings could be extended to dense retrieval in general to help a wide variety of applications.
Moreover, it could be especially beneficial in real applications where the corpus size is much larger than benchmark datasets.


\section*{Limitations}
This work focuses on phrase retrieval, where the training-inference discrepancy might be more significant than other dense retrieval cases, based on the DensePhrases~\cite{lee2021learning} framework.
We plan to explore other dense retrieval methods in the future.
We use open-domain question answering as the main benchmark to show the effectiveness of the proposed method but expect a wide application to other knowledge-intensive tasks.
\section*{Acknowledgements}
The authors appreciate Alon Albalak, Sungdong Kim, Sewon Min, Wenhao Yu, and anonymous reviewers for proofreading the manuscript and giving constructive feedback.
The research presented in this work was funded by Meta AI. The views expressed are those of the authors and do not reflect the official policy or position of the funding agency.

\newpage

\bibliography{ms}
\bibliographystyle{acl_natbib}


\end{document}